\documentclass[11pt,a4paper]{article}

\usepackage{authblk}
\usepackage{array,makecell}
\usepackage{times}
\usepackage{latexsym}
\usepackage{graphics}
\usepackage{color, colortbl}
\usepackage{multirow}
\usepackage[T1]{fontenc}
\usepackage[utf8]{inputenc}
\usepackage{hyperref}
\usepackage[hyphenbreaks]{breakurl}
\usepackage[hyperref]{naaclhlt2019}

\aclfinalcopy % Uncomment this line for the final submission

\usepackage{boldline}

\title{A Silver Standard Corpus of Human Phenotype-Gene Relations}

\author{Diana Sousa\thanks{\ \ dfsousa@lasige.di.fc.ul.pt}\ \ } 
\author{Andre Lamurias}
\author{Francisco M. Couto}

\affil{LASIGE, Faculdade de Ciências, Universidade de Lisboa, Portugal}

\date{}

\begin{document}
\maketitle

% ---------------------------------> ABSTRACT 
\begin{abstract}
 % alamurias: apaguei RE e NER pois as siglas n são usadas
  Human phenotype-gene relations are fundamental to fully understand the origin of some phenotypic abnormalities and their associated diseases. Biomedical literature is the most comprehensive source of these relations, however, we need Relation Extraction tools to automatically recognize them. Most of these tools require an annotated corpus and to the best of our knowledge, there is no corpus available annotated with human phenotype-gene relations. This paper presents the Phenotype-Gene Relations (PGR) corpus, a silver standard corpus of human phenotype and gene annotations and their relations. The corpus consists of 1712 abstracts, 5676 human phenotype annotations, 13835 gene annotations, and 4283 relations\footnote{Query 1, corresponds to the \textit{10/12/2018} release of PGR}. We generated this corpus using Named-Entity Recognition tools, whose results were partially evaluated by eight curators, obtaining a precision of 87.01\%. By using the corpus we were able to obtain promising results with two state-of-the-art deep learning tools, namely 78.05\% of precision. The PGR corpus was made publicly available to the research community.\footnote{\url{https://github.com/lasigeBioTM/PGR}}
 
\end{abstract}

% ---------------------------------> INTRODUCTION
\section{Introduction}

%The volume of unstructured textual information available both on the World Wide Web and in institutional document repositories widely surpasses the ability of analysis by a researcher even if you restring it to a clear-cut topic. If a researcher is interested in a specific subject, searching for articles on that subject and retrieving relevant information is a time-consuming task.

Automatic extraction of relations between entities mentioned in literature is essential to obtain knowledge that was already available but required considerable manual effort and time to retrieve. Recently, biomedical relation extraction has gained momentum in several text-mining applications, such as event extraction and slot-filling \citep{REVIEW}. Some of the commonly extracted biomedical relations are protein-protein interactions \citep{PROTEIN-PROTEIN}, drug-drug interactions \citep{BOLSTM} and disease-gene relationships \citep{DISEASE-GENE}. 

There are a few worth mention systems regarding biomedical Relation Extraction (RE) \citep{N18-1080}, and that specifically focus on the extraction of phenotype-gene relations regarding different species types like plants \citep{PLANTS} and humans \citep{HUMANS}. The main problem that these systems face is a lack of specific high quality annotated corpora (gold standard corpus), mostly because this task requires not only a considerable amount of manual effort but also specific expertise that is not widely available. A solution to these limitations is to generate the corpus in a fully automated manner (silver standard corpus).

Connecting human phenotypes to genes helps us to understand the origin of some phenotypic abnormalities and their associated diseases. To extract human phenotype-gene relations, both entities, human phenotypes and genes have to be recognized. With genes, as a result of lexical features being relatively regular, many systems can successfully identify them in text \citep{BANNER}. Even though Named-Entity Recognition (NER) research has significantly improved in the last years, human phenotype identification is still a complex task, only tackled by a handful of systems \citep{IHP}.

To generate a silver standard for phenotype-gene relation extraction, we used a pipeline that performs: i) NER to recognize genes and human phenotype entities; ii) RE to classify a relation between human phenotype and gene entities. First, we gathered abstracts using the PubMed API with manually defined keywords, %namely \textit{gene name}, 
namely each gene name, \textit{homo sapiens}, and \textit{disease}. Then we used the Minimal Named-Entity Recognizer (MER) tool \citep{MER} to extract gene mentions in the abstracts and the Identifying Human Phenotypes (IHP) tool \citep{IHP} to extract human phenotype mentions. At last, we used a gold standard relations file, provided by the Human Phenotype Ontology (HPO), to classify the relations obtained by co-occurrence in the same sentence as \textit{Known} or \textit{Unknown}.

To the best of our knowledge, there is no corpus available specific to human phenotype-gene relations. This work, overcame this issue by creating a large and versatile silver standard corpus. To assess the quality of the Phenotype-Gene Relations (PGR) corpus, eight curators manually evaluated a subset of PGR. We obtained highly promising results, for example 87.18\% in precision. Finally, we evaluated the impact of using the corpus on two deep learning RE systems, obtaining 69.23\% (BO-LSTM) and 78.05\% (BioBERT) in precision. 

% ---------------------------------> PGR CORPUS
\section{PGR Corpus}

The HPO is responsible for providing a standardized vocabulary of phenotypic abnormalities encountered in human diseases \citep{HPO}. The developers of the HPO also made available a file that links these phenotypic abnormalities to genes. These phenotype-gene relations are regularly extracted from texts in Online Mendelian Inheritance in Man (OMIM) and  Orphanet (ORPHA) databases, where all phenotype terms associated with any disease that is related with a gene are assigned to that gene in the relations file. In this work, we used the relations file created by HPO as a gold standard for human phenotype-gene relations.

We started by retrieving abstracts from PubMed, using the genes involved in phenotype-gene relations and \textit{homo sapiens} as keywords, and the Entrez Programming Utilities (E-utilities) web service (\url{https://www.ncbi.nlm.nih.gov/books/NBK25501/}), retrieving one abstract per gene (Query 1). 

Later, we added the keyword \textit{disease} and filter for abstracts in English (Query 2)\footnote{Query 2, corresponds to the \textit{11/03/2019} release of PGR}.  Query 2 represents a more focused search of the type of abstracts to retrieve, such as abstracts regarding diseases, their associated phenotypes and genes.

For each gene, we opted for the most recent abstract (Query 1) and the two most recent abstracts (Query 2).

% EXAMPLE 1

%\vspace{0.2cm}

%\textbf{Example 1.} PubMed query result example.
%\begin{itemize}

%\vspace{-0.2cm}

%\item\textbf{Keywords:} \textit{NF2} and \textit{homo sapiens}
%\vspace{-0.3cm}
%\item\textbf{Abstract Identifier:} 30194202
%\vspace{-0.3cm}
%\item\textbf{Abstract Title:} Demographical Profile and Spectrum of Multiple Malignancies in Children and Adults with Neurocutaneous Disorders.

%\end{itemize}

% END EXAMPLE 1

We opted by searching per gene and not human phenotype or the combination of both terms because this approach was the one that retrieved abstracts with the higher number of gene and human phenotype annotations, in the following NER and RE phases.
We removed the abstracts that did not check the conditions of being written in English, with a correct XML format and content. The final number of abstracts was 1712 for Query 1 and 2657 for Query 2 as presented in Table \ref{table:results}.
Then we proceeded to use the MER tool \citep{MER} for the annotation of the genes and the IHP framework \citep{IHP} for the annotation of human phenotype terms.

% TABLE 1

\begin{table*}
\renewcommand\arraystretch{1.3}
\footnotesize
\centering
\newcolumntype{C}{ >{\centering\arraybackslash} m{1.3cm} }
\newcolumntype{D}{ >{\centering\arraybackslash} m{1.5cm} }
\newcolumntype{E}{ >{\centering\arraybackslash} m{2cm} }
\begin{tabular}{E|C|D|C|C|C|C}

\hline
%\rowcolor{lightgray}
%\multicolumn{3}{c|}{\textbf{Initial Number}} & \multicolumn{6}{c}{\textbf{Final Number}} \\
%\hline

%\multirow{2}{*}{\textbf{Abstracts}} & \multicolumn{2}{c|}{\textbf{Annotations}} & \multirow{2}{*}{\textbf{Abstracts}} & \multicolumn{2}{c|}{\textbf{Annotations}} & \multicolumn{3}{c}{\textbf{Relations}} \\
\multirow{2}{*}{\textbf{Query}} & \multirow{2}{*}{\textbf{Abstracts}} & \multicolumn{2}{c|}{\textbf{Annotations}} & \multicolumn{3}{c}{\textbf{Relations}} \\

\cline{3-7}
& & \textbf{Phenotype} & \textbf{Gene} & \textbf{Known} & \textbf{Unknown} & \textbf{Total} \\

\hline
\textbf{1} (10/12/2018) & 1712 & 5676 & 13835 & 1510 & 2773 & 4283 \\

\hline
\textbf{2} (11/03/2019) & 2657 & 9553 & 23786 & 2480 & 5483 & 7963 \\
	
\hline
\end{tabular}
\caption{The final number of abstracts retrieved, number of phenotype and gene annotations extracted and the number of known, unknown and total of relations extracted between phenotype and genes, for Query 1 and 2.} 
\label{table:results}
\end{table*}

% END TABLE 1

% -----------------> GENE EXTRACTION
\subsection{Gene Extraction}

MER is a dictionary-based NER tool which given any lexicon or ontology (e.g., an OWL file) and an input text returns a list of recognized entities, their location, and links to their respective classes. 

To annotate genes with MER we need to provide a file of gene names and their respective identifiers. To this goal, we used a list created by the HUGO Gene Nomenclature Committee (HGNC) at the European Bioinformatics Institute (\url{http://www.genenames.org/}). The HGNC is responsible for approving unique symbols and names for human loci, including protein-coding genes, ncRNA genes, and pseudogenes, with the goal of promoting clear scientific communication. Considering that we intended not only to map the genes to their names but also their Entrez Gene (\url{www.ncbi.nlm.nih.gov/gene/}) identifiers, we used the API from MyGene (\url{http://mygene.info/}) with the keyword \textit{human} in species. The MyGene API provides several gene characteristics, including the confidence score for several possible genes that match the query. For this work, we chose the Entrez Gene identifier with a higher confidence score. 

After corresponding all gene names to their respective identifiers, we were left with three genes that did not have identifiers (\textit{CXorf36}, \textit{OR4Q2}, and \textit{SCYGR9}). For the first two genes (\textit{CXorf36} and \textit{OR4Q2}), a simple search in Entrez Gene allowed us to match them to their identifiers. For the last gene (\textit{SCYGR9}) we were not able to find an Entrez Gene identifier, so we used the HGNC identifier for that gene instead. We opted to use the Entrez Gene identifiers because of their widespread use in the biomedical research field. 

To the original gene list, we added gene synonyms using a synonyms list file provided by \url{https://github.com/macarthur-lab/gene_lists} (expanding the original list almost 3-fold). These synonyms were matched to their identifiers and filtered according to their length to exclude one character length synonyms and avoid a fair amount of false positives. The number of genes in the original gene list was 19194, and by including their synonyms that number increased to 56670, representing a total gain of 37476 genes.

At last, we identified some missed gene annotations that were caught using regular expressions. These missed gene annotations were next to forward/back slash and dashes characters (Example 1).

% EXAMPLE 2

\vspace{0.2cm}

\textbf{Example 1.} Missed gene annotation because of forward slash.
\begin{itemize}

\vspace{-0.2cm}

\item\textbf{Gene:} \textit{BAX}
\vspace{-0.3cm}
\item\textbf{Gene Identifier:} 581
\vspace{-0.3cm}
\item\textbf{Abstract Identifier:} 30273005
\vspace{-0.3cm}
\item\textbf{Sentence:} According to the morphological observations and DNA fragmentation assay, the MPS compound induced apoptosis in both cell lines, and also cause a significant increase in the expression of \textbf{Bax}/Bcl-2.

\end{itemize}

% END EXAMPLE 2

% -----------------> PHENOTYPE EXTRACTION
\subsection{Phenotype Extraction}

IHP is a Machine Learning-based  NER tool, specifically created to recognize HPO entities in unstructured text. It uses Stanford CoreNLP \citep{Manning2014} for text processing and applies Conditional Random Fields trained with a rich feature set, combined with hand-crafted validation rules and a dictionary to improve the recognition of phenotypes.

To use the IHP system we chose to update the HPO ontology for the most recent version\footnote{\textit{09/10/2018} release}. The annotations that originated from the IHP system were matched to their HPO identifier. There was a total of 7478 annotations for Query 1 and 10973 annotations for Query 2 that did not match any HPO identifier. We put aside these annotations to be confirmed or discarded manually as some of them are incorrectly identified entities but others are parts of adjacent entities that can be combined for an accurate annotation.

We did not use the MER system for phenotype extraction mainly because a more efficient tool for this task was available without the limitations of a dictionary-based NER tool for complex terms as phenotypes are.

% -----------------> RELATION EXTRACTION
\subsection{Relation Extraction}

After filtering abstracts that did not have annotations of both types, gene and phenotype, we gathered a total of 1712 abstracts for Query 1 and 2656 abstracts for Query 2 as presented in Table \ref{table:results}. The abstracts retrieved by Query 1 were not specific enough for human phenotype-gene relations and therefore about half of them did not contained entities from both types, which we addressed in Query 2, increasing from about 2.5 relations per abstract to about 3.0 relations per abstract.

Using a distant supervision approach, with the HPO file that links phenotypic abnormalities to genes, we were able to classify a relation with \textit{Known} or \textit{Unknown}. For this end, we extract pairs of entities, of gene and human phenotype, by co-occurrence in the same sentence (Example 2). The final number of both \textit{Known} and \textit{Unknown} annotations is also presented in Table \ref{table:results}. 

% EXAMPLE 3

\vspace{0.2cm}

\textbf{Example 2.} Relation extraction.
\begin{itemize}

\vspace{-0.2cm}

\item\textbf{Abstract Identifier:} 23669344
\vspace{-0.3cm}
\item\textbf{Sentence:} A homozygous mutation of \textbf{SERPINB6}, a gene encoding an intracellular protease inhibitor, has recently been associated with post-lingual, autosomal-recessive, nonsyndromic \textbf{hearing loss} in humans (DFNB91).
\vspace{-0.3cm}
\item\textbf{Gene:} \textit{SERPINB6}
\vspace{-0.3cm}
\item\textbf{Gene Identifier:} 5269
\vspace{-0.3cm}
\item\textbf{Phenotype:} \textit{hearing loss}
\vspace{-0.3cm}
\item\textbf{Phenotype Identifier:} HP\_0000365
\vspace{-0.3cm}
\item\textbf{Relation:} \textbf{Known}

\end{itemize}

% END EXAMPLE 3

% ---------------------------------> EVALUATION
\section{Evaluation}

To evaluate the quality of the classifier, we randomly selected 260 relations from Query 1 to be reviewed by eight curators (50 relations each, with an overlap of 20 relations). All researchers work in the areas of Biology and Biochemistry. These curators had to evaluate the correctness of the classifier by attributing to each sentence one of the following options: \textit{C} (correct), \textit{I} (incorrect) or \textit{U} (uncertain). The \textit{U} option was given to identify cases of ambiguity and possible errors in the NER phase. We classified as a true positive (TP) a \textit{Known} relation that was marked \textit{C} by the curator, a false positive (FP) as a \textit{Known} relation marked \textit{I}, a false negative (FN) as a \textit{Unknown} relation marked \textit{I} and a true negative (TN) as a \textit{Unknown} relation marked \textit{C}.

% TABLE 2

\begin{table*}
\renewcommand\arraystretch{1.3}
\footnotesize
\centering
\newcolumntype{C}{ >{\centering\arraybackslash} m{1.1cm} }
\newcolumntype{D}{ >{\centering\arraybackslash} m{1cm} }
\newcolumntype{T}{ >{\centering\arraybackslash} m{1.3cm} }
\newcolumntype{E}{ >{\centering\arraybackslash} m{1.5cm} }

\begin{tabular}{D|T|E|E|E|E|T|D|E}

\hline
%\rowcolor{lightgray}
\multicolumn{2}{c|}{\textbf{Relations}} & \multicolumn{4}{c|}{\textbf{Marked Relations}} & \multicolumn{3}{c}{\textbf{Metrics}} \\

\hline
\textbf{Known} & \textbf{Unknown} & \textbf{True Positive} & \textbf{False Negative} & \textbf{False Positive} & \textbf{True Negative} & \textbf{Precision} & \textbf{Recall} & \textbf{F-Measure}  \\

\hline
77 & 143 & 67 & 86 & 10 & 57 & 0.8701 & 0.4379 & 0.5826 \\
	
\hline
\end{tabular}
\caption{The \textit{Known} and \textit{Unknown} number of relations selected, the number of true positives, false negatives, false positives and true negatives, and the evaluation metrics for the \textit{Known} relations.} 
\label{table:relations}
\end{table*}

% END TABLE 2

% -----------------> STATE-OF-THE-ART APPLICATIONS

\subsection{State-of-the-art Applications}

The PGR corpus was applied to two state-of-the-art systems that were compared against a co-occurrence (or all-true) baseline method. 

% --------> BO-LSTM APPLICATION

\subsubsection{BO-LSTM Application}

The BO-LSTM system \citep{BOLSTM} is a deep learning system that is used to extract and classify relations via long short-term memory networks along biomedical ontologies. This system was initially created to detect and classify drug-drug interactions and later adapted to detect other types of relations between entities like human phenotype-gene relations. It takes advantage of domain-specific ontologies, like the HPO and the Gene Ontology (GO) \citep{GO}. The BO-LSTM system represents each entity as the sequence of its ancestors in their respective ontology. %It is important to point out that the developers of this system matched each gene to their most informative gene ontology concept, prioritizing by evidence code (e.g., Inferred from Experiment).

% --------> BioBERT APPLICATION

\subsubsection{BioBERT Application}

The BioBERT system \citep{BIOBERT} is a pre-trained biomedical language representation model for biomedical text mining based on the BERT \citep{BERT} architecture. Trained on large-scale biomedical corpora, this system is able to perform diverse biomedical text mining tasks, namely NER, RE and Question Answering (QA). The novelty of the architecture is that these systems (BioBERT and BERT) are designed to pre-train deep bidirectional representations by jointly conditioning on both left and right context in all layers. These feature allows easy adaption to several tasks without loss in performance.

% ---------------------------------> RESULTS AND DISCUSSION
\section{Results and Discussion}

The final results are presented in Table \ref{table:relations}. The inter-curator agreement score, calculated from a total of 20 relations classified by eight curators, was 87.58\%. Besides the fact that there were a few incorrectly extracted relations due to errors in the NER phase, that were discarded, the inter-curator agreement is not higher due to the complexity of the sentences where the relations between entities were identified. Even with highly knowledgeable curators in the fields of Biology and Biochemistry, most of them expressed difficulties in deciding which mark to choose on complex sentences that did not necessarily imply a relation between the identified entities (Example 3). 

% EXAMPLE 4

\vspace{0.2cm}

\textbf{Example 3.} Relation marked with \textit{U} (Uncertain).
\begin{itemize}

\vspace{-0.2cm}

\item\textbf{Abstract Identifier:} 27666346
\vspace{-0.3cm}
\item\textbf{Sentence:} \textbf{FRMD4A} antibodies were used to probe 78 paraffin-embedded specimens of \textbf{tongue squamous cell carcinoma} and 15 normal tongue tissues, which served as controls.
\vspace{-0.3cm}
\item\textbf{Mark:} \textbf{\textit{U}}
\end{itemize}

% END EXAMPLE 4

The precision obtained from the test-set (about 6\% of the total of relations), was 87.01\%. Although we cannot state that this test-set is representative of the overall data-set, this is a solid evidence of the effectiveness of our pipeline to automate RE corpus creation, especially between human phenotype and genes, and other domains if a gold standard relations file is provided. Our lower recall is mostly due to incorrectly retrieved human phenotype annotations by IHP, that can be manually confirmed in a future optimized version of the PGR corpus, as some of them are parts of adjacent entities that can be combined for an accurate annotation. 

% -----------------> IMPACT ON DEEP LEARNING
\subsection{Impact on Deep Learning}

For BioBERT we used the available pre-trained weights for training and testing of RE model on our corpus. The results of BO-LSTM and BioBERT in the test-set are presented in Table \ref{table:bolstm_biobert}. We also measured the performance of the co-occurrence (i.e. assuming all-true) baseline method. This baseline method assumes that all relations in the test-set are \textit{Known} and therefore the recall is 100\%. These results are comparable to the ones obtained from the evaluation stage by the curators, and show the applicability of our corpus. 

 BioBERT significantly outperforms BO-LSTM in all metrics proving that is indeed a viable language representation model for biomedical text mining. Even though the recall for both systems is relatively low, the purpose of this work was mainly to extract correct relations between entities to facilitate Machine Learning (ML), which was achieved by obtaining the precision of 69.23\% (BO-LSTM) and 78.95\% (BioBERT).
 
The most relevant metric for a silver standard corpus, directed towards ML tools, is precision. ML tools depend on correct examples to create effective models that can detect new cases, afterwards, being able to deal with small amounts of noise in the assigned labels.

% TABLE 3 

\begin{table}
\renewcommand\arraystretch{1.3}
\footnotesize
\centering
\newcolumntype{C}{ >{\centering\arraybackslash} m{1.2cm} }
\newcolumntype{D}{ >{\centering\arraybackslash} m{1.5cm} }
\newcolumntype{E}{ >{\centering\arraybackslash} m{2cm} }

\begin{tabular}{E|C|C|D}

\hline
\textbf{Method} & \textbf{Precision} & \textbf{Recall} & \textbf{F-Measure}\\

\hline
\textbf{Co-occurrence} & 0.3500 & 1.0000 & 0.5185 \\

\hline
\textbf{BO-LSTM} & 0.6923 & 0.4200 & 0.5228 \\

\hline
\textbf{BioBERT} & 0.7895 & 0.5844 & 0.6716 \\

\hline
\end{tabular}
\caption{Precision, recall and F-measure of the co-occurrence baseline, BO-LSTM and BioBERT.} 
\label{table:bolstm_biobert}
\end{table}

% END TABLE 3

% ---------------------------------> CONCLUSIONS
\section{Conclusions}

This paper showed that our pipeline is a feasible way of generating a silver standard human phenotype-gene relation corpus. The pipeline required the application of two NER tools, and the availability of a list of known relations. We manually evaluated the corpus using eight curators obtaining a 87.01\% precision with an inter-agreement of 87.58\%. We also measured the impact of using the corpus in state-of-the-art deep learning RE systems, namely BO-LSTM and BioBERT. The results were promising with 69.23\%, and 78.95\% in precision, respectively. We believe that our pipeline and silver standard corpus will be a highly useful contribution to overcome the lack of gold standards.

Future work includes manually correcting the human phenotype annotations that did not match any HPO identifier, with the potential of expanding the number of human phenotype annotations almost 2-fold and increasing the overall recall. Also, we intend to expand the corpus by identifying more missed gene annotations using pattern matching, which is possible due to our approach being fully automated. Another possibility is the expansion of the test-set for a more accurate capture of the variance in the corpus. For example, we can select a subset of annotated documents in which two curators could work to grasp the complexity of manually annotating some of these abstracts. Further, we intend to use semantic similarity to validate the human phenotype-gene relations. Semantic similarity has been used to compare different types of biomedical entities \citep{SSM}, and is a measure of closeness based on their biological role. For example, if the \textit{BRCA1} gene is semantically similar to the \textit{BRAF} gene and the \textit{BRCA1} has an established relation with the \textit{tumor} phenotype, it could be possible to infer that \textit{BRAF} gene also has a relation with the \textit{tumor} phenotype, even if that is not evident by the training data. 
Finally, the effect of different NER systems applied to the pipeline should be studied.

% ---------------------------------> ACKNOWLEDGMENTS
\section*{Acknowledgments}

We acknowledge the help of Márcia Barros, Joana Matos, Rita Sousa, Ana Margarida Vasconcelos, Maria Teresa Cunha and Sofia Jesus in the curating phase.

This work was supported by FCT through funding of DeST: Deep Semantic Tagger project, ref. PTDC/CCI-BIO/28685/2017 (\url{http://dest.rd.ciencias.ulisboa.pt/}), and LASIGE Research Unit, ref. UID/CEC/00408/2019.

\bibliography{naaclhlt2019}
\bibliographystyle{acl_natbib}

\end{document}